\documentclass[conference]{IEEEtran}
\IEEEoverridecommandlockouts
\usepackage{mathrsfs}
\usepackage{amsmath}
\usepackage{amssymb}
\usepackage{amsthm}
\usepackage{multirow,makecell}
\usepackage[table]{xcolor}
\usepackage{setspace}
\usepackage{stfloats}
\usepackage{graphicx}

\ifCLASSINFOpdf \else \fi

\usepackage{algorithm}
\usepackage{algorithmic}

\usepackage{cite}

\begin{document}
\title{Spectrum Sharing between UAV-based Wireless Mesh
Networks and Ground Networks}
\author{Zhiqing Wei\IEEEauthorrefmark{1},
Zijun Guo\IEEEauthorrefmark{1},
Zhiyong Feng\IEEEauthorrefmark{1},
Jialin Zhu\IEEEauthorrefmark{1},
Caijun Zhong\IEEEauthorrefmark{2},
Qihui Wu\IEEEauthorrefmark{3},
Huici Wu\IEEEauthorrefmark{1}\\
\IEEEauthorblockA{\IEEEauthorrefmark{1}Key Laboratory of Universal Wireless Communications, Ministry of Education,\\
\IEEEauthorrefmark{1}Beijing University of Posts and Telecommunications, Beijing, China\\
\IEEEauthorrefmark{2}Institute of Information and Communication Engineering, Zhejiang University, Hangzhou, China\\
\IEEEauthorrefmark{3}College of Electronic and Information Engineering,\\
\IEEEauthorrefmark{3}Nanjing University of Aeronautics and Astronautics, Nanjing, China\\
Email: \IEEEauthorrefmark{1}\{weizhiqing, zijunguo, fengzy, jialinzhu, dailywu\}@bupt.edu.cn}
\IEEEauthorrefmark{2}caijunzhong@zju.edu.cn,
\IEEEauthorrefmark{3}wuqihui2014@sina.com}

\maketitle

\begin{abstract}
The unmanned aerial vehicle (UAV)-based
wireless mesh networks
can economically provide wireless services
for the areas with disasters.
However,
the capacity of air-to-air communications
is limited due to the multi-hop transmissions.
In this paper,
the spectrum sharing between UAV-based wireless mesh
networks and ground networks is studied to improve
the capacity of the UAV networks.
Considering the distribution of UAVs as a
three-dimensional (3D) homogeneous Poisson point process (PPP)
within a vertical range,
the stochastic geometry is applied to
analyze the impact of
the height of UAVs, the transmit power of UAVs,
the density of UAVs and the vertical range, etc.,
on the coverage probability of ground network user
and UAV network user, respectively.
The optimal height of UAVs is numerically achieved in
maximizing the capacity of UAV networks
with the constraint of the coverage
probability of ground network user.
This paper provides a basic guideline for the deployment of
UAV-based wireless mesh networks.
\end{abstract}

\IEEEpeerreviewmaketitle

\section{Introduction}

Since unmanned aerial vehicles (UAVs)
have flexible maneuverability and
large coverage,
the UAV-mounted base stations (BSs) are widely
applied to provide ubiquitous
wireless connections \cite{UAV_Survey}.
The UAV-mounted BSs relieve
the mismatch between the diverse traffic load and
the fixed infrastructures \cite{ UAV_aerial_BS}.
For example, the demand for mobile and flexible wireless connections
is urgent in the areas with traffic congestions or concerts.  UAV-mounted BSs can be deployed in this scenario
to offload the cellular traffic to the UAVs \cite{wu}.
In the areas with disasters, the ground
infrastructures are destroyed and the UAV-mounted BSs
can be deployed to
provide communication services to the
rescue persons and vehicles on ground \cite{UAV_Ground_Cooperation, UAV_VEL}.


In the UAV-mounted BS system,
multiple small UAVs can provide more economical wireless coverage
than a single large UAV \cite{UAV_Multi}.
Li \emph{et al.} in \cite{UAV_TWO}
developed a two-UAV relaying system
to extend the communication range
of UAV networks.
They have verified the feasibility
of realizing multi-UAV communications.
Chand \emph{et al.} in \cite{UAV_MESH_1}
designed a UAV-based wireless mesh network
for the scenarios of disaster management and
military environment.
The project loon established by Alphabet Inc.
aims to provide Internet access to
remote areas using balloons
which form an aerial mesh
network \cite{Google_loon}.
In \cite{ICCC}, we have realized
the UAV-based wireless mesh network,
where multiple UAVs provide wireless
coverage to the users on ground.
Meanwhile, the UAVs form an aerial ad hoc network.
With one UAV accessing the Internet via the gateway,
all the users on ground can
access the Internet.

According to Gupta and Kumar's theory \cite{Kumar},
the per-node capacity of ad hoc networks
is a decreasing function of the number of hops.
For the aerial tier in the UAV-based mesh network,
the multi-hop transmissions
bring severe
capacity shortage problem for each UAV.
Spectrum sharing is an effective technology in
improving the capacity of wireless
networks via enhancing the
spectrum utilization.
Fortunately,
the UAV networks and
ground networks such as cellular networks are spatially separated,
which creates a unique opportunity
for the spectrum sharing between them \cite{cog_survey}.
Zhang \emph{et al.} in \cite{cog_zhangwei} studied the
spectrum sharing between the drone small cell networks and the
cellular networks.
Sboui \emph{et al.} in \cite{cog_ee} optimized the
transmit power to
maximize the energy efficiency when a UAV shares the
spectrum of primary users.
Lyu \emph{et al.} in \cite{cog_zengyong} designed
the orthogonal spectrum sharing between UAV and ground BS.
Yoshikawa \emph{et al.} in \cite{cog_3d} studied the spectrum
sharing between UAVs and radar systems.
Huang \emph{et al.} in \cite{cog_routing} designed the
routing schemes
for the aerial cognitive radio networks.

Although many prior works have studied the
issue of spectrum sharing between UAVs and
other wireless systems,
to the best of the authors' knowledge,
very few studies have considered the
issue of spectrum sharing between the UAV-based
wireless mesh networks and
the ground networks,
such as cellular networks.
Motivated by this,
in this paper, the spectrum sharing is studied in the
air-to-air communications of UAVs to improve the
capacity of UAV networks.
Considering the distribution of UAVs as a
three-dimensional (3D) homogeneous Poisson point process (PPP),
stochastic geometry is applied to
analyze the coverage probability of UAV network users
and ground network users.
As a result, the optimal height of UAVs can be
found with the constraint of
the coverage probability of ground network users.

The remainder of this paper is organized as follows.
In Section II, the system model is introduced.
Section III analyzes the performance of spectrum sharing
of UAV-ground networks using stochastic geometry.
The simulation results are provided in Section IV.
Finally, we summarize this paper in Section V.

\section{System Model}

\begin{figure}
\centering
\includegraphics[width=0.48\textwidth]{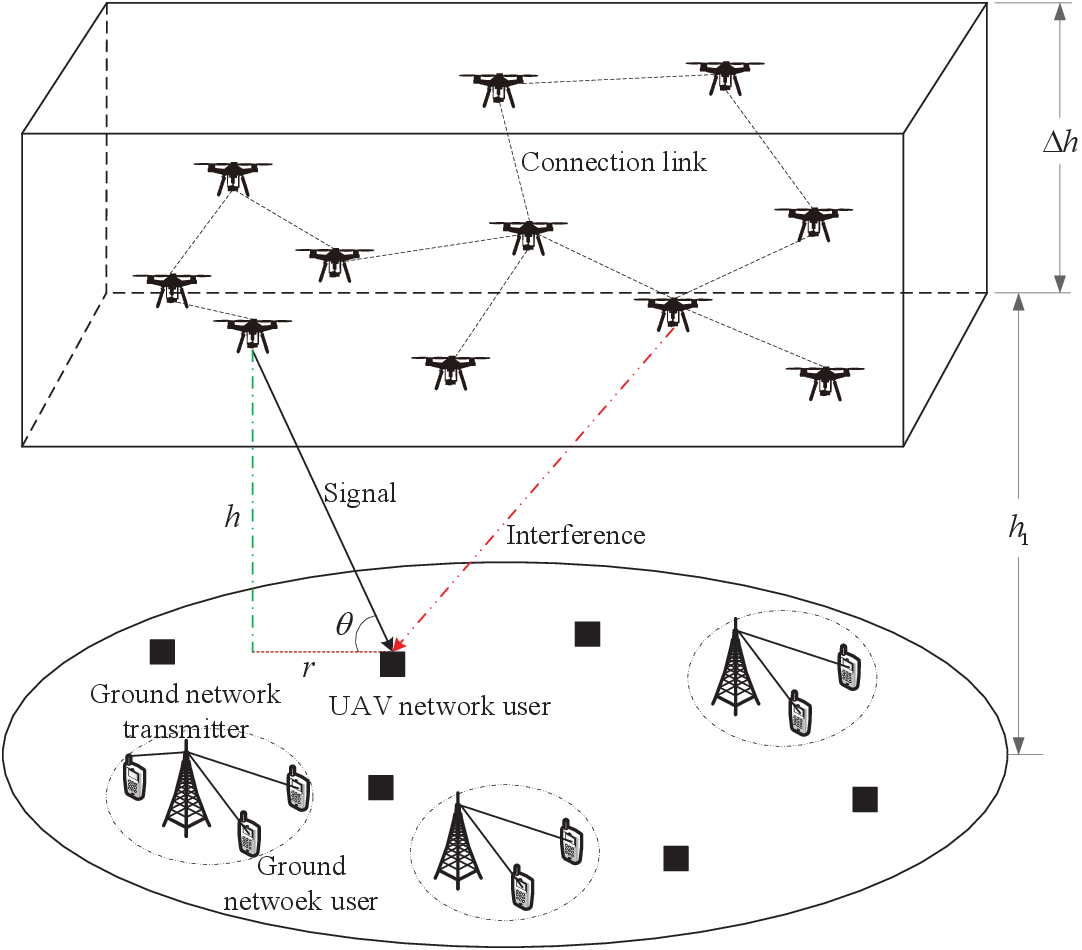}
\caption{System model of UAV-ground spectrum sharing.}
\label{fig_system_model_3D}
\end{figure}

When multiple UAVs provide wireless
services to the
users on ground,
the UAVs should form a wireless mesh network
to improve the coverage of the multiple UAVs.
In this scenario,
each UAV acts as an aerial BS
and the multiple UAVs form an aerial ad hoc network.
With one UAV connecting to the gateway via backhaul link,
all the UAVs can provide Internet connections for the
UAV network users.
In the aerial ad hoc networks,
multi-hop transmissions are required
to forward data from sources to destinations.
Since multi-hop transmissions
bring severe capacity shortage problem for each UAV,
the communications among UAVs share the
spectrum of the ground networks to improve the network
capacity.
While the spectrum of the
air-to-ground communications is
different from the spectrum of
the ground networks to avoid
the severe interference between them.

\subsection{Network model}

As illustrated in Fig. \ref{fig_system_model_3D},
the distribution of the
transmitters (TXs) of ground network,
such as the BSs of cellular network\footnote{In this paper, the general ground networks including, but not limited to cellular network, are considered.},
follows a two-dimensional (2D)
homogeneous PPP ${\Phi _{d}}$ with
density ${\lambda _d}$.
The distribution of UAVs follows
a 3D
homogeneous PPP ${\Phi _{u}}$ with density
${\lambda_u}$.
The minimum and maximum heights of UAVs
are $h_1$ and $h_1 + \Delta h$, respectively.
$\Delta h$ is defined as the vertical range of the UAVs.
In the UAV network,
the Aloha protocol is applied
as the medium access
control (MAC) protocol.

\subsection{Channel model}

Both path-loss and small scale fading are considered.
The path-loss exponent of the ground-to-ground link is
$\alpha_d$.
The path-loss exponent of the air-to-ground link is
$\alpha_u$.
The power gain of small scale fading is an
exponential distributed random variable with unit mean.
Additive white Gaussian noise is considered with mean
zero and variance $N$.
The transmit power of the UAV and the TX of ground network
are $P_u$ and $P_d$, respectively.
Both the line-of-sight (LoS) and non-line-of-sight (NLoS)
propagations in the air-to-ground communications are considered.
The received signal strength at the
UAV network user can be expressed as \cite{UAV_D2D, UAV_Channel}
\begin{equation}\label{Eq:1}
{P_{r,g}} = \left\{ {\begin{array}{*{20}{c}}
{{P_u}{{\left| {{d_{ug}}} \right|}^{ - {\alpha _u}}}}&{{\rm{LoS}}}\\
{\eta {P_u}{{\left| {{d_{ug}}} \right|}^{ - {\alpha _u}}}}&{{\rm{NLoS}}}
\end{array}} \right.,
\end{equation}
where ${{d_{ug}}}$ is the distance between the
UAV and the UAV network user. $\eta$
is an attenuation
factor because of the NLoS propagation \cite{UAV_D2D}.
The probability of LoS propagation
is as follows \cite{UAV_Channel}.
\begin{equation}
{P_{\rm{LoS}}} = \frac{1}{{1 + C\exp \left( { - B(\theta  - C)} \right)}},
\end{equation}
where $B$ and $C$ are environmental dependent constants.
$\theta$ is the elevation angle.
As illustrated in Fig. \ref{fig_system_model_3D},
with $h$ being the height of a UAV and $r$ being the distance
between the projection of the UAV on ground and
the UAV network user,
the value of $\theta$ is
\begin{equation}
\theta  = \frac{{180}}{\pi }\arctan ( {\frac{h}{r}} ).
\end{equation}

\section{Spectrum Sharing of UAV-Ground Networks}

\newcounter{mytempeqncnt}
\begin{figure*}[ht]
\normalsize
\begin{equation}\label{eq_DG}
\begin{aligned}
{L_{I_{gu}^c}}( {\frac{{\beta {d_0}^{{\alpha _d}}}}{{{P_d}}}} ) = \exp ( { - \frac{{2\lambda_d {\pi ^2}{{\left( \beta  \right)}^{2/{\alpha _d}}}{d_0}^2}}{{{\alpha _d}\sin \left( {2\pi /{\alpha _d}} \right)}}})
\end{aligned}
\end{equation}

\begin{equation}\label{eq_L_LOS}
\begin{aligned}
{L_{{I_{u,LOS}}}}( {\frac{{\beta d_0^{{\alpha _d}}}}{{{P_d}}}} ) & = \exp ( { - {\lambda _u}\int\limits_V {( {1 - \frac{1}{{1 + \frac{{\beta d_0^{{\alpha _d}}}}{{{P_d}}}{P_u}{P_{LOS}}x_i^{ - {\alpha _u}}}}} )} dx} )\\
& = \exp ( { - {\lambda _u}\int_{h1}^{h2} {\int_0^{2\pi } {\int_0^\infty  {( {1 - \frac{1}{{1 + \frac{{\beta d_0^{{\alpha _d}}}}{{{P_d}}}{P_u}{{(\sqrt {{r^2} + {z^2}} )}^{ - {\alpha _u}}}\frac{1}{{1 + C\exp ( { - B(\frac{{180}}{\pi }\arctan (z/r) - C)} )}}}}} )} } } rdrd\phi dz} )\\
& = \exp ( { - 2\pi {\lambda _u}{H_1}( {\beta ,{d_0},h,{\alpha _d},{\alpha _u}} )} ).
\end{aligned}
\end{equation}

\begin{equation}\label{eq_L_NLOS}
\begin{aligned}
{L_{{I_{u,{\text{NLoS}}}}}}( {\frac{{\beta d_0^{{\alpha _d}}}}{{{P_d}}}} ) & = \exp ( { - 2\pi {\lambda _u}\int_{h1}^{h2} {\int_0^\infty  {( {1 - \frac{1}{{1 + \frac{{\beta d_0^{{\alpha _d}}}}{{{P_d}}}{P_u}\eta {{(\sqrt {{r^2} + {z^2}} )}^{ - {\alpha _u}}}(1 - \frac{1}{{1 + C\exp ( { - B(\frac{{180}}{\pi }\arctan (z/r) - C)} )}})}}} )} } rdrdz} )\\
& = \exp ( { - 2\pi {\lambda _u}{H_2}( {\beta ,{d_0},h,{\alpha _d},{\alpha _u}} )} ).
\end{aligned}
\end{equation}

\begin{equation}\label{eq_h1}
{H_1}( {\beta ,{d_0},h,{\alpha _d},{\alpha _u}} ) = \int_{h1}^{h2} {\int_0^\infty  {( {1 - \frac{1}{{1 + \frac{{\beta d_0^{{\alpha _d}}}}{{{P_d}}}{P_u}{{(\sqrt {{r^2} + {z^2}} )}^{ - {\alpha _u}}}\frac{1}{{1 + C\exp ( { - B(\frac{{180}}{\pi }\arctan (z/r) - C)} )}}}}} )} } rdrdz.
\end{equation}

\begin{equation}\label{eq_h2}
{H_2}( {\beta ,{d_0},h,{\alpha _d},{\alpha _u}} ) = \int_{h1}^{h2} {\int_0^\infty  {( {1 - \frac{1}{{1 + \frac{{\beta d_0^{{\alpha _d}}}}{{{P_d}}}{P_u}\eta {{(\sqrt {{r^2} + {z^2}} )}^{ - {\alpha _u}}}(1 - \frac{1}{{1 + C\exp ( { - B(\frac{{180}}{\pi }\arctan (z/r) - C)} )}})}}} )} } rdrdz.
\end{equation}
\hrulefill
\end{figure*}

With the spectrum sharing between the
UAV network and the ground network,
we derive the coverage probabilities of
ground network user and UAV network user
respectively.

The coverage probability is defined as the probability of
successful communication.
Define $\beta$ as the threshold of
signal-to-interference-plus-noise ratio (SINR)
at the receiver for successfully communication.
The coverage probability is the probability $P(\gamma > \beta )$
with $\gamma$ being the SINR of the receiver.

\subsection{The coverage probability of ground network user}
Define ${\gamma_{gu}}=\frac{{{P_d}d_0^{ - {\alpha _d}}{g_0}}}{{I_{gu}^c + {I_u} + N}} $ as the received SINR of a typical ground network user at the origin $\{ \bf{0}\}$,
where $g_0$ is the power gain of small scale fading.
$d_0$ is the distance between the typical ground network user and its associated TX.
${I_{gu}^c}$ and ${{I_u}}$ are the interference generated by the TXs of ground network and the UAVs, respectively.
\begin{equation}
I_{gu}^c = \sum\limits_{{d_i} \in {\Phi _d}\backslash \{ \bf{0}\} } {{P_d}} d_i^{ - {\alpha _d}}{g_i},
\end{equation}
\begin{equation}
\begin{aligned}
{I_u} & = {I_{u,LoS}} + {I_{u,{\rm{NLoS}}}}\\
& = \sum\limits_{{x_i} \in {\Phi _u}} {{P_{{\rm{LoS}}}}{P_u}} x_i^{ - {\alpha _d}}{g_i} + \sum\limits_{{x_i} \in {\Phi _u}} {\left( {1 - {P_{{\rm{LoS}}}}} \right)\eta {P_u}} x_i^{ - {\alpha _d}}{g_i},
\end{aligned}
\end{equation}
where $g_i$ is the small scale fading gain of the interference link.
$d_i$ is the distance between the $i$th TX of ground network and the typical ground network user.
$x_i$ is the distance between the $i$th UAV and the typical ground network user.

With the definition of the coverage probability,
the coverage probability of a typical ground network user is
\begin{equation}
\begin{aligned}
{P_1} &= P( {\gamma_{gu}} > \beta),\\
& \mathop  = \limits^{(a)} \exp ( { - \frac{{\beta d_0^{{\alpha _d}}( {I_{gu}^c + {I_u} + N} )}}{{{P_d}}}} )\\
& = \exp ( { - \frac{{\beta d_0^{{\alpha _d}}I_{gu}^c}}{{{P_d}}}} )\exp ( { - \frac{{\beta d_0^{{\alpha _d}}{I_u}}}{{{P_d}}}} )\exp ( { - \frac{{\beta d_0^{{\alpha _d}}N}}{{{P_d}}}} )\\
& = {L_{I_{gu}^c}}( {\frac{{\beta d_0^{{\alpha _d}}}}{{{P_d}}}} ){L_{{I_u}}}( {\frac{{\beta d_0^{{\alpha _d}}}}{{{P_d}}}} )\exp ( { - \frac{{\beta d_0^{{\alpha _d}}N}}{{{P_d}}}} ),
\end{aligned}
\end{equation}
where $(a)$ is obtained from the exponential distribution of $g_0$.
${L_A}(*)$ is the Laplace transform of the random variable $A$.

With the considered path-loss model (\ref{Eq:1}),
$I_u$ can be re-expressed as

\begin{equation}
I_u={I_{u,{\rm{LoS}}}}+{I_{u,{\rm{NLoS}}}}.
\end{equation}

The Laplace function of $I_u$ then can be expressed as
\begin{equation}
 {L_{{I_u}}}( {\frac{{\beta d_0^{{\alpha _d}}}}{{{P_d}}}} ) = {L_{{I_{u,{\rm{LoS}}}}}}( {\frac{{\beta d_0^{{\alpha _d}}}}{{{P_d}}}} ){L_{{I_{u,N{\rm{LoS}}}}}}( {\frac{{\beta d_0^{{\alpha _d}}}}{{{P_d}}}} ).
\end{equation}

Since $g_i$ is a random variable independent of the point process ${\Phi _u}$,
we have
\begin{equation}
\begin{aligned}
& {L_{{I_{u,{\rm{LoS}}}}}}( {\frac{{\beta d_0^{{\alpha _d}}}}{{{P_d}}}} ) = {E_{{I_{u,{\rm{LoS}}}}}}[ {\exp ( {\frac{{\beta d_0^{{\alpha _d}}}}{{{P_d}}}{I_{u,{\rm{LoS}}}}} )} ]\\
& = {E_{{g_i},{\Phi _u}}}[ {\prod\limits_{{x_i} \in {\Phi _u}\backslash \{ \bf{0}\} } {\exp ( {\frac{{\beta d_0^{{\alpha _d}}}}{{{P_d}}}{g_i}{P_u}{P_{{\rm{LoS}}}}x_i^{ - {\alpha _u}}} )} } ]\\
& = {E_{{\Phi _u}}}[ {\prod\limits_{{x_i} \in {\Phi _u}\backslash \{ \bf{0}\} } {{E_{{g_i}}}[\exp ( {\frac{{\beta d_0^{{\alpha _d}}}}{{{P_d}}}{P_u}{P_{{\rm{LoS}}}}x_i^{ - {\alpha _u}}} )]} } ]\\
& = {E_{{\Phi _u}}}[ {\prod\limits_{{x_i} \in {\Phi _u}\backslash \{ \bf{0}\} } {\frac{1}{{1 + \frac{{\beta d_0^{{\alpha _d}}}}{{{P_d}}}{P_u}{P_{{\rm{LoS}}}}x_i^{ - {\alpha _u}}}}} } ],
\end{aligned}
\end{equation}

and
\begin{equation}
\begin{aligned}
& {L_{{I_{u,{\rm{NLoS}}}}}}( {\frac{{\beta d_0^{{\alpha _d}}}}{{{P_d}}}} )\\
& = {E_{{\Phi _u}}}[ {\prod\limits_{{x_i} \in {\Phi _u}\backslash \{ \bf{0}\} } {\frac{1}{{1 + \frac{{\beta d_0^{{\alpha _d}}}}{{{P_d}}}{P_u}(1 - {P_{{\rm{LoS}}}})\eta x_i^{ - {\alpha _u}}}}} } ].
\end{aligned}
\end{equation}

Hence, ${L_{{I_u}}}( {\frac{{\beta d_0^{{\alpha _d}}}}{{{P_d}}}} )$ is derived as follows.
\begin{equation}
\begin{aligned}
& {L_{{I_u}}}( {\frac{{\beta d_0^{{\alpha _d}}}}{{{P_d}}}} ) = {E_{{\Phi _u}}}[ {\prod\limits_{{x_i} \in {\Phi _u}} {\frac{1}{{1 + \frac{{\beta d_0^{{\alpha _d}}}}{{{P_d}}}{P_u}{P_{{\rm{LoS}}}}x_i^{ - {\alpha _u}}}}} } ] \\
&\times {E_{{\Phi _u}}}[ {\prod\limits_{{x_i} \in {\Phi _u}} {\frac{1}{{1 + \frac{{\beta d_0^{{\alpha _d}}}}{{{P_d}}}{P_u}(1 - {P_{{\rm{LoS}}}})\eta x_i^{ - {\alpha _u}}}}} } ].
\end{aligned}
\end{equation}

Applying the probability generating function of PPP \cite{cog_zhangwei}
\begin{equation}\label{eq_pgf}
E( {\prod\limits_{{x_i} \in \Phi } {f( x )} } ) = \exp ( { - {\lambda _d}\int\limits_V {[ {1 - f( x )} ]} dx} ),
\end{equation}
then
${L_{I_{gu}^c}}( {\frac{{\beta d_0^{{\alpha _d}}}}{{{P_d}}}} )$
can be derived as (\ref{eq_DG}) \cite{123}.
${L_{{I_u}}}( {\frac{{\beta d_0^{{\alpha _d}}}}{{{P_d}}}} )$
can be derived using (\ref{eq_L_LOS}) and (\ref{eq_L_NLOS}),
where ${H_1}( {\beta ,{d_0},h,{\alpha _d},{\alpha _u}} )$ and ${H_2}( {\beta ,{d_0},h,{\alpha _d},{\alpha _u}} )$
are provided in (\ref{eq_h1}) and (\ref{eq_h2}), respectively.

\begin{figure*}[ht]
\normalsize
\begin{equation}\label{eq_uiasduia}
\begin{aligned}
{L_{I_{u,{\rm{LoS}}}^c}}( {\frac{{\beta x_0^{{\alpha _u}}}}{{{P_u}}}} ) & = \exp ( { - {\lambda _u}\int\limits_V {( {1 - \frac{1}{{1 + \beta x_0^{{\alpha _u}}{P_{{\rm{LoS}}}}x_i^{ - {\alpha _u}}}}} )} dx} )\\
& = \exp ( { - {\lambda _u}\int_{h1}^{h2} {\int_0^{2\pi } {\int_0^\infty  {( {1 - \frac{1}{{1 + \beta x_0^{{\alpha _u}}{{(\sqrt {{r^2} + {z^2}} )}^{ - {\alpha _u}}}\frac{1}{{1 + C\exp ( { - B(\frac{{180}}{\pi }\arctan (z/r) - C)} )}}}}} )} } } rdrd\phi dz} )\\
& = \exp ( { - 2\pi {\lambda _u}{H_3}( {\beta ,{x_0},h,{\alpha _u}} )} ).
\end{aligned}
\end{equation}

\begin{equation}\label{eq_ugsg}
\begin{aligned}
{L_{I_{u,{\rm{NLoS}}}^c}}( {\frac{{\beta x_0^{{\alpha _u}}}}{{{P_u}}}} ) & = \exp ( { - 2\pi {\lambda _u}\int_{h1}^{h2} {\int_0^\infty  {( {1 - \frac{1}{{1 + \beta x_0^{{\alpha _u}}\eta {{(\sqrt {{r^2} + {z^2}} )}^{ - {\alpha _u}}}(1 - \frac{1}{{1 + C\exp ( { - B(\frac{{180}}{\pi }arctan(z/r) - C)} )}})}}} )} } rdrdz} )\\
& = \exp ( { - 2\pi {\lambda _u}{H_4}( {\beta ,{x_0},h,{\alpha _u}} )} ).
\end{aligned}
\end{equation}

\begin{equation}\label{eq_h3}
{H_3}( {\beta ,{x_0},h,{\alpha _u}} ) = \int_{h1}^{h2} {\int_0^\infty  {( {1 - \frac{1}{{1 + \beta x_0^{{\alpha _u}}{{(\sqrt {{r^2} + {z^2}} )}^{ - {\alpha _u}}}\frac{1}{{1 + C\exp ( { - B(\frac{{180}}{\pi }\arctan (z/r) - C)} )}}}}} )} } rdrdz.
\end{equation}

\begin{equation}\label{eq_h4}
{H_4}( {\beta ,{x_0},h,{\alpha _u}} ) = \int_{h1}^{h2} {\int_0^\infty  {( {1 - \frac{1}{{1 + \beta x_0^{{\alpha _u}}\eta {{(\sqrt {{r^2} + {z^2}} )}^{ - {\alpha _u}}}(1 - \frac{1}{{1 + C\exp ( { - B(\frac{{180}}{\pi }\arctan (z/r) - C)} )}})}}} )} } rdrdz.
\end{equation}
\hrulefill \vspace*{4pt}
\end{figure*}

\subsection{The coverage probability of UAV network user}

The coverage probability of a typical UAV network user is
defined as
\begin{equation}
{P_2} = P( {\gamma_{uu} > \beta } ),
\end{equation}
where $\gamma_{uu}$ is the received SINR of the UAV network user and $\beta$
is the SINR threshold\footnote{Note that although we use the same parameter $\beta$ for the SINR threshold, the values of $\beta$
for ground network and UAV network can be different.}.
With the considered path-loss model (\ref{Eq:1}),
${P_2}$ can be expressed as
\begin{equation}\label{eq_coverage_probability_UAV_user}
\begin{aligned}
P_2 & = {P_{{\text{LoS}}}}P( {\frac{{{P_u}x_0^{ - {\alpha _u}}{g_i}}}{{I_u^c}} > \beta } ) + {P_{{\text{NLoS}}}}P( {\frac{{\eta {P_u}x_0^{ - {\alpha _u}}{g_i}}}{{I_u^c}} > \beta } )\\
& = {P_{{\text{LoS}}}}\exp ( { - \frac{{\beta x_0^{{\alpha _u}}I_u^c}}{{{P_u}}}} ) + (1 - {P_{{\text{LoS}}}})\exp ( { - \frac{{\beta x_0^{{\alpha _u}}I_u^c}}{{\eta {P_u}}}} )\\
& = {P_{{\text{LoS}}}}{L_{I_u^c}}( { - \frac{{\beta x_0^{{\alpha _u}}}}{{{P_u}}}} ) + (1 - {P_{{\text{LoS}}}}){L_{I_u^c}}( { - \frac{{\beta x_0^{{\alpha _u}}}}{{\eta {P_u}}}} ),
\end{aligned}
\end{equation}
where $x_0$ is the distance between the typical UAV user and its associated UAV.
The term $I_u^c$ is the received interference from UAVs
and we have
\begin{equation}\label{eq_yuasd}
\begin{aligned}
I_u^c & = I_{_{u,{\text{LoS}}}}^c + I_{_{u,{\text{NLoS}}}}^c\\
&  = \sum\limits_{{x_i} \in {\Phi _u}\backslash \{ \bf{0}\} } {{P_{{\text{LoS}}}}{P_u}} x_i^{ - {\alpha _u}}{g_i} + \\
& \sum\limits_{{x_i} \in {\Phi _u}\backslash \{ \bf{0}\} } {( {1 - {P_{{\text{LoS}}}}} )\eta {P_u}} x_i^{ - {\alpha _u}}{g_i}.
\end{aligned}
\end{equation}

Similar to the derivation of the
coverage probability of ground network user, we have
\begin{equation}\label{eq_yuiasyuda}
\begin{aligned}
{L_{I_u^c}}( {\frac{{\beta x_0^{{\alpha _u}}}}{{{P_u}}}} ) & = {L_{I_{u,{\text{LoS}}}^c}}( {\frac{{\beta x_0^{{\alpha _u}}}}{{{P_u}}}} ){L_{I_{u,{\text{NLoS}}}^c}}( {\frac{{\beta x_0^{{\alpha _u}}}}{{{P_u}}}} )\\
& = {E_{{\Phi _u}}}[ {\prod\limits_{{x_i} \in {\Phi _u}\backslash \{ \bf{0}\} } {\frac{1}{{1 + \beta x_0^{{\alpha _u}}{P_{{\text{LoS}}}}x_i^{ - {\alpha _u}}}}} } ] \times  \hfill \\
& {E_{{\Phi _u}}}[ {\prod\limits_{{x_i} \in {\Phi _u}\backslash \{ \bf{0}\} } {\frac{1}{{1 + \beta x_0^{{\alpha _u}}(1 - {P_{{\text{LoS}}}})\eta x_i^{ - {\alpha _u}}}}} } ].
\end{aligned}
\end{equation}

The ${L_{I_{u,{\text{LoS}}}^c}}( {\frac{{\beta x_0^{{\alpha _u}}}}{{{P_u}}}} )$
and ${L_{I_{u,{\text{NLoS}}}^c}}( {\frac{{\beta x_0^{{\alpha _u}}}}{{{P_u}}}} )$ can be derived
in (\ref{eq_uiasduia}) and (\ref{eq_ugsg}), where
${H_3}( {\beta ,{x_0},h,{\alpha _u}} )$ and ${H_4}( {\beta ,{x_0},h,{\alpha _u}} )$ are provided in (\ref{eq_h3}) and (\ref{eq_h4}), respectively.

\section{Numerical Results and Analysis}

This section provides the numerical results of the
coverage probabilities of UAV network user and
ground network user.
Besides, the transmission capacity
of UAV network is defined and
maximized.
The parameters in the simulations are summarized in Table 1.

\begin{table}[!t]
 \caption{Simulation parameters}
 \begin{center}
 \begin{tabular}{l|l}
 \hline
 \hline
    {Parameter} & {Value} \\
  \hline
  $P_u$ & 5 W \\
  $P_d$ & 0.1 W \\
  $\alpha_u$ & 3 \\
  $\alpha_d$ & 4 \\
  $B$ and $C$ & 0.136 and 11.95\\
  $\beta$ & 0.1 \\
  $\eta$ & 0.1\\
  $\lambda_u$ & $10^{-4}$ per square meter\\
  $\lambda_d$ & $10^{-3}$ per square meter\\
  $d_0$ & 10 m\\
  $N$ & $10^{-9}$ W\\
  \hline
  \hline
 \end{tabular}
 \end{center}
\end{table}

\subsection{The coverage probability of ground network user}

\begin{figure}[!t]
\centering
\includegraphics[width=0.48\textwidth]{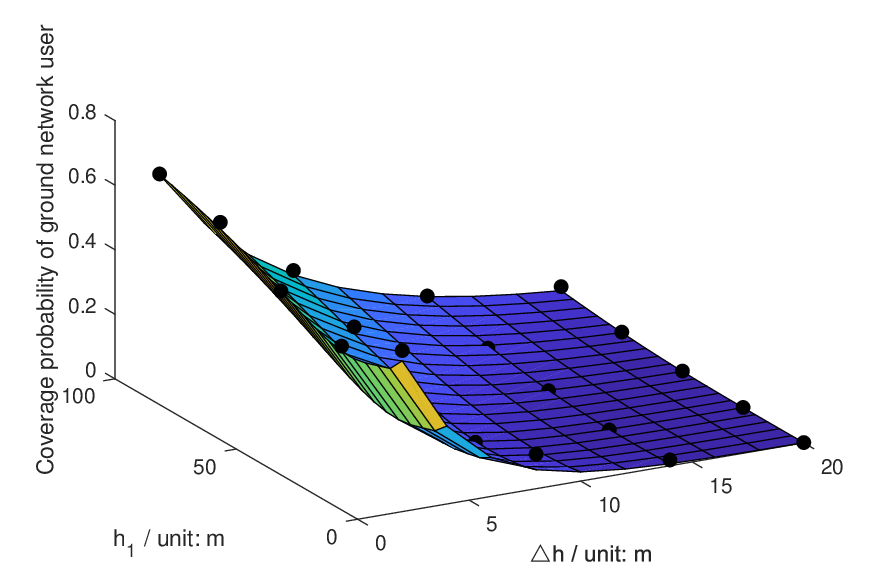}
\caption{The coverage probability of ground network user versus $h_1$ and $\Delta h$.}
\label{fig_d2dv1}
\end{figure}

\begin{figure}[!t]
\centering
\includegraphics[width=0.48\textwidth]{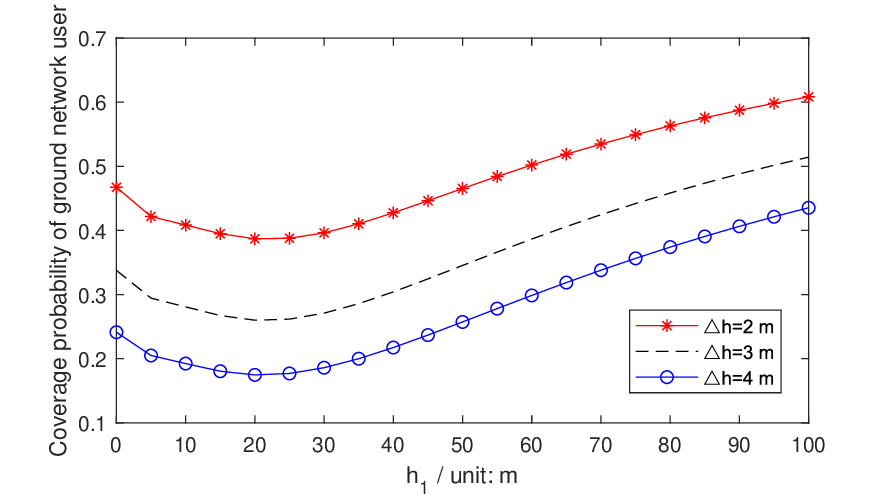}
\caption{The relation between the coverage probability of ground network user and $h_1$ with different values of $\Delta h$.}
\label{fig_d2d}
\end{figure}

The coverage probability of a typical ground network user,
namely, $P_1$ is illustrated in Fig. \ref{fig_d2dv1}
as a function of $h_1$ and $\Delta h$.
The 20-point Monte Carlo simulation results are
provided in Fig. \ref{fig_d2dv1}.
Each point undergoes 1000 times Monte Carlo simulations.
It is verified that the theoretical results,
namely, the surface fits well with the points.
Notice that when $h_1$ is large, for example, when
$h_1$ is close to 100 m, $P_1$ is large.
This is due to the fact that
when the UAVs are high above the ground,
the interference from UAVs to the typical ground
network user is small, which will increase the value of $P_1$.
When $\Delta h$ is increasing,
$P_1$ is decreasing because the probability of LoS propagation
from UAVs to the ground network user is increasing.
This discovery is also observed in Fig. \ref{fig_d2d},
which depicts the relation between $P_1$
and $h_1$ with different values of $\Delta h$.
In Fig. \ref{fig_d2d},
when $h_1$ is increasing from 0,
$P_1$ is firstly decreasing because the probability of LoS propagation
between the UAV and the ground network user is increasing.
When $h_1$ exceeds a threshold,
$P_1$ is increasing with
the increase of $h_1$ because the propagation path between
the UAV and the ground network user becomes
long in this case, which will decrease
the interference from UAVs to ground network user.

\subsection{The coverage probability of UAV network user}

\begin{figure}[!t]
\centering
\includegraphics[width=0.48\textwidth]{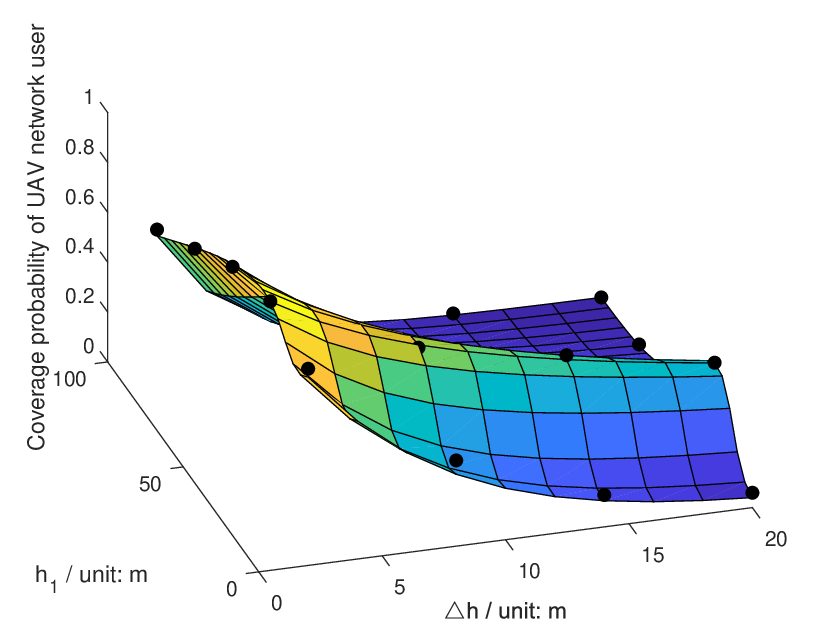}
\caption{The coverage probability of UAV network user versus $h_1$ and $\Delta h$.}
\label{fig_du3d_omni}
\end{figure}

The coverage probability of UAV network user, namely, $P_2$
is illustrated
in Fig. \ref{fig_du3d_omni}.
The 20-point Monte Carlo simulation results are provided in Fig. \ref{fig_du3d_omni}.
Each point undergoes 1000 times Monte Carlo simulations.
Notice that the theoretical results,
namely, the surface fits well with the points.
The $P_2$ fluctuates with the increase of $h_1$.
For each $\Delta h$, there exists an optimal $h_1$
to maximize $P_2$.
Besides, with the increase of $\Delta h$,
$P_2$ is decreasing because the signal link is long.
A critical observation is that the
when $\Delta h \to 0$,
namely, when UAVs are distributed in
2D plane, $P_2$
has maximum value in Fig. \ref{fig_du3d_omni}.

\subsection{The optimal height of UAVs}

The definition of transmission capacity (TC) in \cite{cog_zhangwei}
is applied to verify the performance of UAV network, which is
as follows \cite{cog_zhangwei}.
\begin{equation}
T_u = {\lambda _u}P(\gamma_{uu} > \beta )\log (1 + \beta ),
\end{equation}
where $\gamma_{uu}$ is the SINR of the UAV network user and
$T_u$ denotes the TC
of UAV network.
With the constraint of the coverage probability of
ground network user, the optimal height of UAVs, defined as $h_1$,
can be found to maximize the TC of
UAV network.
The optimization model is as
follows.
\begin{equation}\label{eq_optimization}
\begin{aligned}
& \mathop {\max }\limits_{{h_1}} {\kern 3pt} {T_u}\\
& s.t.{\kern 9pt} {P_1} \ge \alpha.
\end{aligned}
\end{equation}

Although the form of (\ref{eq_optimization}) is simple, the object
function and constraint condition are complex.
It is difficult to derive a closed-form solution.
Hence the optimal solution of $h_1$ is derived numerically.

\begin{figure}[!t]
\centering
\includegraphics[width=0.47\textwidth]{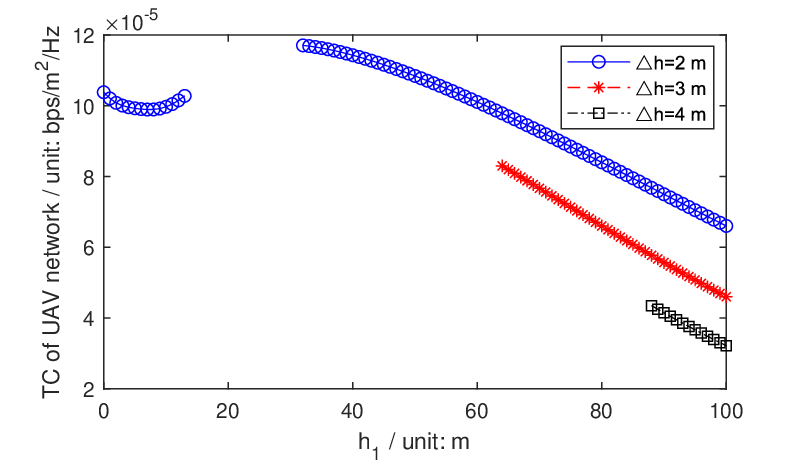}
\caption{The relation between the transmission capacity of UAV network user and $h_1$ with $\alpha = 0.4$ and different values of $\Delta h$.}
\label{fig_du1}
\end{figure}

\begin{figure}[!t]
\centering
\includegraphics[width=0.47\textwidth]{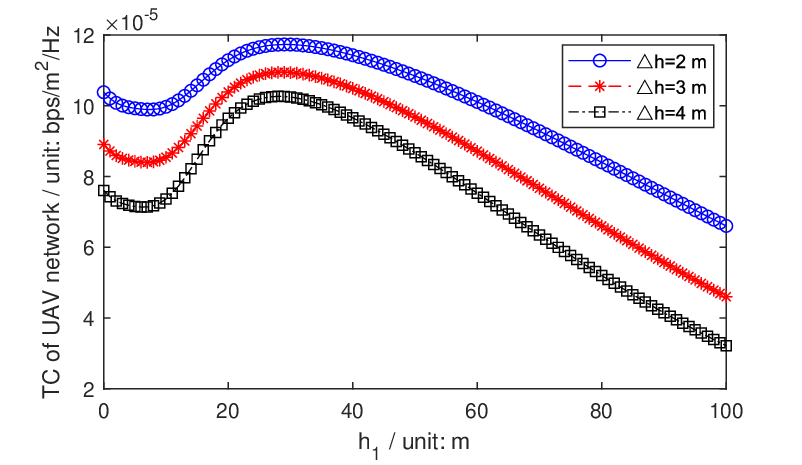}
\caption{The relation between the transmission capacity of UAV network user and $h_1$ with $\alpha = 0.1$ and different values of $\Delta h$.}
\label{fig_du2}
\end{figure}

With $\alpha = 0.4$, the relation between the
TC of UAV network and $h_1$ is illustrated
in Fig. \ref{fig_du1}.
The optimal $h_1$ to maximize the
TC of UAV network can be searched.
It is verified that with the decrease of $\Delta h$,
the TC of UAV network increases.
When $\Delta h \to 0$, namely, the UAVs are distributed in a
2D plane, the TC of UAV network is maximum.
With the constraint $\alpha = 0.4$, there are vacant segments where
the $h_1$ does not
satisfy the constraint of (\ref{eq_optimization}).
However, when the constraint $\alpha = 0.1$,
all the values of $h_1$ in Fig. \ref{fig_du2} are feasible solutions.
In this case, the optimal $h_1$ can still be searched
to maximize the TC of UAV network.

\section{Conclusion}

In this paper,
the spectrum sharing between UAV-based wireless mesh
networks and ground networks is analyzed
using stochastic geometry.
The impact of
the height of UAVs,
the transmit power of UAVs,
the density of UAVs
and the vertical range
on the coverage probability of ground network user
and UAV network user is analyzed.
Then the optimal height of UAVs is achieved to
maximize the transmission capacity of UAV networks.
This paper provides fundamental analysis for the spectrum sharing of
UAV-based wireless mesh networks, which
may motivate the study of spectrum sharing
for more aerial wireless mesh networks.

\section*{Acknowledgment}

This work is supported by National
Natural Science Foundation of
China (No. 61601055, No. 61631003, No. 61525101).

\end{document}